\documentclass[10pt,journal,compsoc]{IEEEtran}
\ifCLASSOPTIONcompsoc
  \usepackage[nocompress]{cite}
\else
  \usepackage{cite}
\fi
\ifCLASSINFOpdf
\else
\fi
\usepackage{times}

\usepackage{soul}
\usepackage{url}
\usepackage[hidelinks]{hyperref}
\usepackage[utf8]{inputenc}
\usepackage[small]{caption}
\usepackage{graphicx}
\usepackage{amsmath}
\usepackage{booktabs}
\usepackage{subfigure}
\usepackage{xcolor}
\usepackage{algorithm}
\usepackage{algorithmic}
\usepackage{multirow}
\usepackage{bm}
\usepackage{amsmath}
\usepackage{amssymb}
\usepackage{amsthm}

\usepackage{cancel}
\usepackage{multicol}
\usepackage{threeparttable}
\usepackage[switch]{lineno}

\begin{document}
\title{Sliceformer: Make Multi-head Attention as Simple as Sorting in Discriminative Tasks}

\author{Shen~Yuan,~       
Hongteng~Xu,~\IEEEmembership{Member,~IEEE}% <-this % stops a space
\IEEEcompsocitemizethanks{
\IEEEcompsocthanksitem Shen Yuan was with the Gaoling School of Artificial Intelligence, Renmin University of China.
\protect\\
E-mail: shenyuan721@ruc.edu.cn
\IEEEcompsocthanksitem Hongteng Xu was the correspondence author of the paper. 
He was with the Gaoling School of Artificial Intelligence, Renmin University of China and Beijing Key Laboratory of Big Data Management and Analysis Methods. 
\protect\\
E-mail: hongtengxu@ruc.edu.cn
}% <-this % stops an unwanted space
\thanks{Manuscript submitted April XX, 2021.}}


\markboth{Journal of \LaTeX\ Class Files,~Vol.~14, No.~8, August~20XX}%
{Shell \MakeLowercase{\textit{et al.}}: Bare Demo of IEEEtran.cls for Computer Society Journals}



\IEEEtitleabstractindextext{%
\begin{abstract}
As one of the most popular neural network modules, Transformer plays a central role in many fundamental deep learning models, e.g., the ViT in computer vision and the BERT and GPT in natural language processing.
The effectiveness of the Transformer is often attributed to its multi-head attention (MHA) mechanism. 
In this study, we discuss the limitations of MHA, including the high computational complexity due to its ``query-key-value'' architecture and the numerical issue caused by its softmax operation. 
Considering the above problems and the recent development tendency of the attention layer, we propose an effective and efficient surrogate of the Transformer, called Sliceformer. 
Our Sliceformer replaces the classic MHA mechanism with an extremely simple ``slicing-sorting'' operation, i.e., projecting inputs linearly to a latent space and sorting them along different feature dimensions (or equivalently, called channels).
For each feature dimension, the sorting operation implicitly generates an implicit attention map with sparse, full-rank, and doubly-stochastic structures.
We consider different implementations of the slicing-sorting operation and analyze their impacts on the Sliceformer. 
We test the Sliceformer in the Long-Range Arena benchmark, image classification, text classification, and molecular property prediction, demonstrating its advantage in computational complexity and universal effectiveness in discriminative tasks. 
Our Sliceformer achieves comparable or better performance with lower memory cost and faster speed than the Transformer and its variants. 
Moreover, the experimental results reveal that applying our Sliceformer can empirically suppress the risk of mode collapse when representing data.
The code is available at \url{https://github.com/SDS-Lab/sliceformer}.
\end{abstract}

}

\maketitle

\IEEEdisplaynontitleabstractindextext

% \fi
\IEEEpeerreviewmaketitle

\section{Introduction}\label{sec:intro}
\IEEEPARstart{T}{ransformer}~\cite{vaswani2017attention} has been dominant in deep learning research for recent years. 
It works as a backbone module in many fundamental models, achieving outstanding performance in various application scenarios. 
Currently, the most successful language models like BERT~\cite{devlin2019bert} and GPT~\cite{brown2020language} are built based on the Transformer or its variants~\cite{child2019generating,dai2019transformer}, which outperforms classic recurrent neural network (RNN) architectures on both effectiveness and efficiency. 
In the field of computer vision, the Vision Transformers (ViTs)~\cite{dosovitskiy2021an,liu2021swin,arnab2021vivit} have achieved better performance in many image recognition and understanding tasks compared to convolutional neural networks (CNNs).
Recently, the Transformer-based models have been designed for the structured data in different applications, including the Informer~\cite{zhang2021informer} for time series broadcasting, the Graphormer~\cite{ying2021transformers} for molecular representation, the Set-Transformer~\cite{lee2019set} and Point-Transformer~\cite{zhao2021point} for point cloud modeling, and so on.

More and more cases show the tendency that the Transformer is becoming an indispensable choice when developing deep learning models.
Note that some work makes attempts to replace the Transformer with some other architectures, including the MLP Mixer for vision tasks~\cite{tolstikhin2021mlp}, the RNN-based competitor of the Transformer~\cite{katharopoulos2020transformers}, the Structured State Space Sequential (S4) Model~\cite{gu2021efficiently} and its simplified variant S5~\cite{smith2022simplified} for modeling extremely-long sequences, and so on. 
Still, they mainly focus on reusing and improving existing models (e.g., MLP, RNN, and State Space model) in specific tasks (e.g., image classification and sequential prediction) rather than designing a new module applicable for general purposes.
As a result, although these models can outperform the Transformer in one or two applications, none are as universally useful as the Transformer.

\begin{table*}[t]
    \caption{A comparison for representative Transformers and our Sliceformer on their attention mechanisms. 
    We show one attention head for each transformer, in which the input $\bm{X}\in\mathbb{R}^{N\times d}$, the value $\bm{V}=\bm{X}\bm{W}_V\in\mathbb{R}^{N\times D}$, the query $\bm{Q}=\bm{X}\bm{W}_Q\in\mathbb{R}^{N\times D}$, and the key $\bm{K}=\bm{X}\bm{W}_K\in\mathbb{R}^{N\times D}$.}
    \label{tab:cmp}
    \centering
    % \small{
    \begin{threeparttable}
    {
    \def\arraystretch{1.5}\tabcolsep=10pt
    \begin{tabular}{l|lll}
    \toprule
    Model & 
    $\text{Attention}(\bm{V};\bm{Q},\bm{K})$ & 
    Complexity & 
    Attention Structure \\
    \midrule
    Transformer~\cite{vaswani2017attention}  & 
    $\text{Softmax}\left(\frac{\bm{Q}\bm{K}^{\top}}{\sqrt{D}}\right)\bm{V}$ & 
    $\mathcal{O}(DN^2)$ & 
    Dense + Row-wisely normalized\\
    SparseTrans~\cite{child2019generating} & 
    $\text{Local2D-Softmax}\left(\frac{\bm{Q}\bm{K}^{\top}}{\sqrt{D}}\right)\bm{V}$ & 
    $\mathcal{O}(DN^{1.5})$ &
    Sparse + Row-wisely normalized\\
    Longformer~\cite{beltagy2020longformer}   & 
    $\text{Local1D-Softmax}\left(\frac{\bm{Q}\bm{K}^{\top}}{\sqrt{D}}\right)\bm{V}$ &  
    $\mathcal{O}(DNL)$ &
    Sparse + Row-wisely normalized\\
    Reformer~\cite{kitaev2020reformer}     & 
    $\text{LSH-Softmax}\left(\frac{\bm{Q}\bm{K}^{\top}}{\sqrt{D}}\right)\bm{V}$ &  
    $\mathcal{O}(DN\log N)$ &
    Sparse + Row-wisely normalized\\
    CosFormer~\cite{zhen2022cosformer}    & 
    $(\bm{Q}_{\cos}\bm{K}_{\cos}^{\top}+\bm{Q}_{\sin}\bm{K}_{\sin}^{\top})\bm{V}$ & 
    $\mathcal{O}(\min\{DE_{QK},NE_{Q}\})$&
    Sparse\\
    Performer~\cite{choromanski2021rethinking}  & 
    $\phi_r(\bm{Q})\phi_r(\bm{K})^{\top}\bm{V}$ & 
    $\mathcal{O}(DNr)$ &
    Low-rank\\
    Linformer~\cite{wang2020linformer}    & 
    $\text{Softmax}\left(\frac{\bm{Q}\psi_r(\bm{K})^{\top}}{\sqrt{D}}\right)\psi_r(\bm{V})$ & 
    $\mathcal{O}(DNr)$ &
    Low-rank + Row-wisely normalized\\
    Sinkformer~\cite{sander2022sinkformers}   & 
    $\text{Sinkhorn}_{K}\left(\frac{\bm{Q}\bm{K}^{\top}}{\sqrt{D}}\right)\bm{V}$ & 
    $\mathcal{O}(KDN^2)$ &
    Dense + Doubly stochastic\\
    \midrule
    {\textbf{Sliceformer}}  & 
    {$\text{Sort}_{\text{col}}(\bm{V})$} & 
    {$\mathcal{O}(DN\log N)$} &
    Full-rank + Sparse + Doubly stochastic\\
    \bottomrule
    \end{tabular}
    }
    \begin{tablenotes}
    \item[1] \footnotesize{``Local1D'' considers $L$ local data in a sequence. ``Local2D'' considers the row-wise and column-wise local data for a sequence zigzagging in the 2D space. ``LSH'' denotes Locality-sensitive Hashing.}
    \item[2] \footnotesize{$\phi_r: \mathbb{R}^{D}\mapsto\mathbb{R}^r$, and $\phi_r(\bm{Q}),\phi_r(\bm{K})\in\mathbb{R}^{N\times r}$; $\psi_r: \mathbb{R}^{N}\mapsto\mathbb{R}^r$, and $\psi_r(\bm{K}),\psi_r(\bm{V})\in\mathbb{R}^{r\times D}$.}
    \item[3] $\bm{K}_{\cos}=\text{diag}(\{\cos\frac{\pi i}{2M}\}_{i=1}^{N})\text{ReLU}(\bm{K})$, $\bm{K}_{\sin}=\text{diag}(\{\sin\frac{\pi i}{2M}\}_{i=1}^{N})\text{ReLU}(\bm{K})$. So are $\bm{Q}_{\cos}$ and $\bm{Q}_{\sin}$. 
    $E_{QK}$ is the number of nonzero elements in $\bm{Q}_{\cos}\bm{K}_{\cos}^{\top}$. $E_Q$ is the number of nonzero elements in $\bm{Q}_{\cos}$.
    \item[4] ``$\text{Sinkhorn}_{K}$'' means applying $K$-step Sinkhorn iterations.
    \item[5] Note that, our Sliceformer does not need the ``multi-head'' architecture because of the simplicity of sorting.
    \end{tablenotes}
    \end{threeparttable}
% }
\end{table*}

Although without strict theoretical support, the effectiveness of the Transformer is often attributed to the multi-head attention (MHA) mechanism~\cite{vaswani2017attention} behind it.
This empirical but dominant opinion impacts the design and modification of the Transformer significantly, which, in our opinion, might have restricted the development of new model architectures to some degree.
As shown in Table~\ref{tab:cmp}, many variants of Transformer have been proposed to $i)$ improve the efficiency of MHA (e.g., designing sparse or low-rank attention maps~\cite{child2019generating,kitaev2020reformer,wang2020linformer}), $ii)$ enhance the interpretability of MHA (e.g., revisiting attention maps through the lens of kernel theory~\cite{tsai2019transformer,zhen2022cosformer} and optimal transport~\cite{tay2020sparse,sander2022sinkformers}), or $iii)$ impose more side information on attention maps~\cite{dong2021attention,ying2021transformers,ma2022mega}. 
However, these Transformer-driven models still rely on the classic ``query-key-value'' (abbreviately, QKV) architecture of MHA. 
Little attention is paid to studying the necessity of the architecture or, more ambitiously, replacing it with a new surrogate for general purposes.

In this study, we focus on discriminative learning tasks and challenge the architecture of MHA, proposing an extremely simple ``slicing-sorting'' operation and developing a surrogate of the Transformer called Sliceformer. 
In particular, our work is motivated by the MHA's drawbacks and the attention map's possibly desired structures. 
Firstly, we attribute the MHA's numerical issues to the softmax operation and its high complexity to its QKV architecture.
Secondly, the development tendency of the MHA and its variants implies that we shall pursue as many sparse, full-rank, and doubly stochastic attention maps as possible for projected samples. 
Based on the analysis above, we propose the ``slicing-sorting'' operation, which projects samples linearly to a latent space and sorts them along different feature dimensions (a.k.a. channels). 
Replacing the MHA mechanism of the Transformer with the slicing-sorting operation leads to the proposed Sliceformer.

We analyze the connections and differences between the proposed slicing-sorting operation and the MHA mechanism and discuss its rationality in depth. 
Essentially, the ``slicing-sorting'' operation generates channel-wise attention maps implicitly as permutation matrices for the projected samples.
Therefore, the attention maps are sparse, full-rank, and doubly stochastic matrices, satisfying the desired structures mentioned above.
As shown in Table~\ref{tab:cmp}, different from the classic QKV architecture, our slicing-sorting operation only preserves the linear map from the input $\bm{X}$ to the value matrix $\bm{V}$. 
We do not need the ``multi-head'' structure because concatenating different linear maps is equivalent to directly increasing the columns of $\bm{W}_V$. 
As a result, our Sliceformer has fewer parameters and lower computational complexity than the Transformer and its variants. 
In addition, the sorting step of the Sliceformer can be implemented in ascending or descending order.
To further enhance the diversity of the learned implicit attention maps, we can apply the ascending and descending sorting operations in an interleaving manner for each layer of the Sliceformer and change the frequency of the interleaves across different layers. 

% We analyze the connections and differences between the proposed slicing-sorting operation and the MHA mechanism and discuss its rationality in depth. 
% Specifically, we find that although abandoning the MHA architecture, the slicing-sorting operation achieves sparse and doubly stochastic attention maps by a set of permutation matrices. 

\begin{figure}[t]
  \centering
\includegraphics[height=8cm]{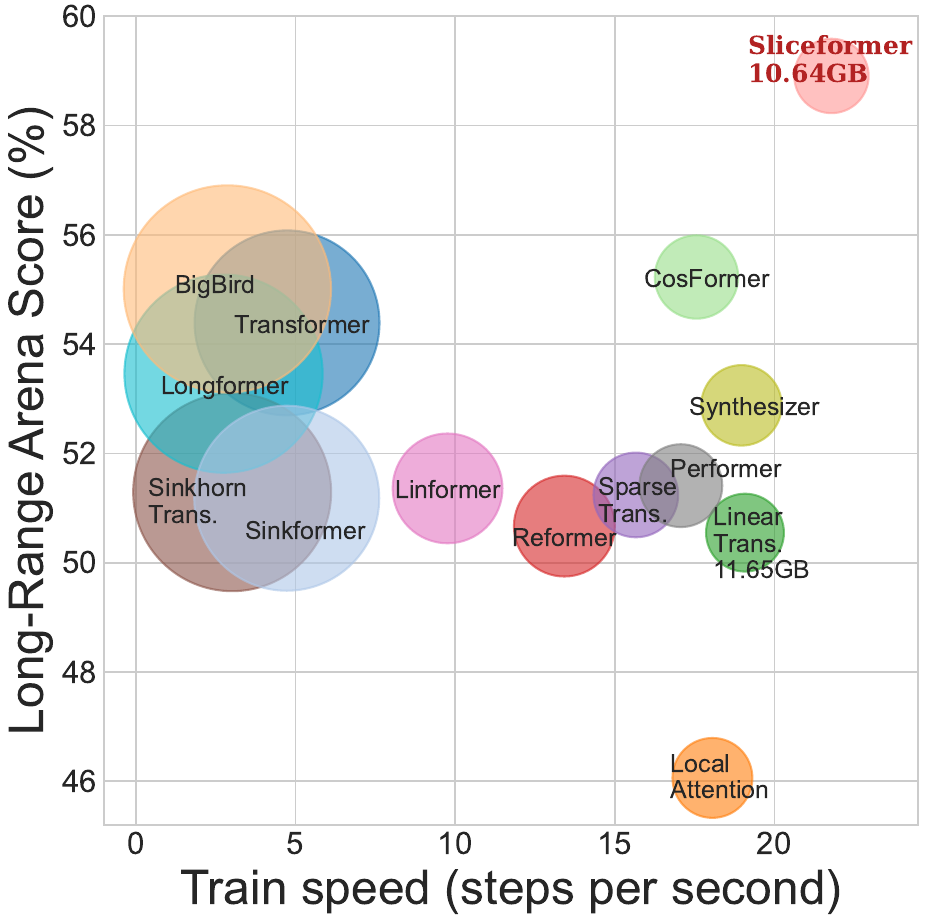}
    \caption{The comparison for various Transformers and our Sliceformer on the LRA benchmark. 
    The length of the sequence is 3K.
    The x-axis corresponds to the number of training steps per second. 
    The y-axis corresponds to the average score (\%) on the LRA benchmark.
    The peak memory usage of each model is represented as the area of the corresponding circle. 
    For a better comparison, the values (GB) of the top-2 models are shown.}
    \label{fig:cmp}
\end{figure}

We test our Sliceformer in the well-known Long-Range Arena (LRA) benchmark, demonstrating its advantages in extremely long sequential modeling. 
In particular, as shown in Fig.~\ref{fig:cmp}, our Sliceformer achieves superior performance with less memory cost and runtime than the Transformer and its variants. 
Ablation studies demonstrate the rationality of our model setting. 
Furthermore, through other discriminative learning tasks, including image classification, text classification, and molecular property prediction, we further demonstrate the universal applicability of our Sliceformer.

% The remainder of this paper is organized as follows:
% Section~\ref{sec:related} provides a detailed literature review and explains the connections and differences between our work and existing methods. 
% Section~\ref{sec:model} introduces the proposed Sliceformer and demonstrates its rationality and advantages. 
% Section~\ref{sec:exp} provides the implementation details of our Sliceformer and experimental results in various discriminative learning tasks. 
% Finally, Section~\ref{sec:con} concludes the paper, discussing the limitations of our model and pointing out our future work. 

\section{Related Work}\label{sec:related}
\subsection{Transformer and Its Applications}
Transformer~\cite{vaswani2017attention} is a powerful sequential model friendly to parallel computing. 
Since it was proposed, Transformer has become the critical module of many large language models, e.g., BERT~\cite{devlin2019bert}, Transformer-XL~\cite{dai2019transformer}, and GPT~\cite{brown2020language}. 
Besides texts, Transformer is also applied to other sequences, e.g., the Music Transformer~\cite{huang2018music} for music modeling, the Informer~\cite{zhang2021informer} for time series broadcasting and the Transformer Hawkes process~\cite{zuo2020transformer} for event sequence prediction. 
For non-sequential data like images, the Vision Transformer (ViT)~\cite{dosovitskiy2021an} and its variants~\cite{liu2021swin,chen2021visformer} take the patches of images as a sequence and extract image representations accordingly, which outperform convolutional neural networks (CNNs) in image classification.
Nowadays, Transformer is being introduced for structured data modeling, e.g., the Graphormer~\cite{ying2021transformers} for molecules, the Set-Transformer~\cite{lee2019set} for point clouds, and the Mesh Transformer~\cite{lin2021mesh} for 3D meshes. 
The more applications the Transformer has, the more significant it is to study its architecture, especially its MHA mechanism. 

\subsection{The Variants of Transformer}
It has been known that the classic MHA mechanism suffers from high computational complexity, poor scalability, and numerical instability for long sequences. 
As shown in Table~\ref{tab:cmp}, many efforts have been made to overcome these issues. 
The SparseTrans in~\cite{child2019generating} and the Longformer in~\cite{beltagy2020longformer} compute local attention maps based on the sub-sequences extracted by sliding windows, which leads to sparse global attention maps.
Some other models sparsify the key and query matrices directly by the locality-sensitive hashing (LSH)~\cite{kitaev2020reformer} or the ReLU operation~\cite{zhen2022cosformer}. 
Besides pursuing sparse attention maps, another technical route is constructing low-rank attention maps. 
The Performer in~\cite{choromanski2021rethinking} reduces the feature dimension (the column number) of the query and key matrices, while the Linformer in~\cite{wang2020linformer} reduces the sample dimension (the row number) of the key and value matrices. 

In addition to simplifying the computation of the attention maps, some work provides new understandings of the attention mechanism. 
The work in~\cite{tsai2019transformer} treats the attention map as a normalized linear kernel and revisits the vanilla Transformer through different kernels.
The Performer~\cite{choromanski2021rethinking} and the CosFormer~\cite{zhen2022cosformer} introduce additional mappings for the query and key matrices and consider their linear kernels in the latent spaces.
Recently, the work~\cite{sander2022sinkformers} reports an interesting phenomenon that the attention map tends to be doubly stochastic during training. 
Accordingly, it implements the attention map as an optimal transport through the Sinkhorn-Knopp algorithm~\cite{sinkhorn1967concerning}. 
Note that although providing these new understandings, these Transformer variants fail to design new model architectures that overcome the MHA's issues.

In recent years, there have been several attempts to replace this mechanism with alternative architectures, such as the works in~\cite{gu2021efficiently,smith2022simplified}. 
They have endeavored to scale to long sequence inputs from the perspective of State Space Models, resulting in a significant improvement in capturing long-range dependencies.
However, these models can only serve as substitutes for Transformer in the sequence modeling tasks and lack the generality of Transformer.

\section{Proposed Sliceformer}\label{sec:model}
\subsection{Motivation and Design Principle}\label{sec:Motivations}
Typically, given an input $\bm{X}\in\mathbb{R}^{N\times d}$, where $N$ indicates the length of a sequence or the size of a sample set and $d$ is the input feature dimension, an attention head first obtains the value, query, and key matrices by linear maps, i.e., $\bm{V}=\bm{X}\bm{W}_V\in\mathbb{R}^{N\times D}$, $\bm{Q}=\bm{X}\bm{W}_Q\in\mathbb{R}^{N\times D}$, and $\bm{K}=\bm{X}\bm{W}_K\in\mathbb{R}^{N\times D}$, and then projects $\bm{V}$ as follows:
\begin{eqnarray}\label{eq:att}
\begin{aligned}
    \text{Att}(\bm{V};\bm{Q},\bm{K}):=\bm{P}(\bm{Q},\bm{K})\bm{V}. 
\end{aligned}
\end{eqnarray}
Here, we take $\bm{V}$ as the input of the head, and $\bm{P}(\bm{Q},\bm{K})\in\mathbb{R}^{N\times N}$ is the attention map parametrized by $\bm{Q}$ and $\bm{K}$. 
The multi-head attention layer applies a group of linear maps, i.e., $\theta=\{\bm{W}_{V,m},\bm{W}_{Q,m},\bm{W}_{K,m}\in\mathbb{R}^{d\times D}\}_{m=1}^{M}$, to construct $M$ attention heads and concatenates their outputs, i.e.,
\begin{eqnarray}\label{eq:mha}
\begin{aligned}
    \text{MHA}_{\theta}(\bm{X}) := \text{Concat}_{\text{col}}(\{\text{Att}(\bm{V}_m;\bm{Q}_m,\bm{K}_m)\}_{m=1}^{M}),
\end{aligned}
\end{eqnarray}
where $\text{MHA}_{\theta}(\bm{X})\in\mathbb{R}^{N\times MD}$, $\text{Concat}_{\text{col}}$ means the column-wise concatenation of the input matrices, and for $m=1,...,M$, we have $\bm{V}_m=\bm{X}\bm{W}_{V,m}$, $\bm{Q}_m=\bm{X}\bm{W}_{Q,m}$, and $\bm{K}_m=\bm{X}\bm{W}_{K,m}$.

The vanilla Transformer implements the attention map based on the softmax operation, i.e.,
\begin{eqnarray}\label{eq:att_softmax}
\begin{aligned}
    \bm{P}(\bm{Q},\bm{K}) = \text{Softmax}\Bigl(\frac{\bm{QK}^{\top}}{\sqrt{D}}\Bigr),
\end{aligned}
\end{eqnarray}
where $\text{Softmax}(\cdot)$ is applied to each row of the matrix $\frac{\bm{QK}^{\top}}{\sqrt{D}}$.
Following this strategy, most existing variants of the Transformer implement their attention maps based on the softmax operation as well, as shown in Table~\ref{tab:cmp}.

\begin{figure}[t]
    \centering
    \includegraphics[height=3.5cm]{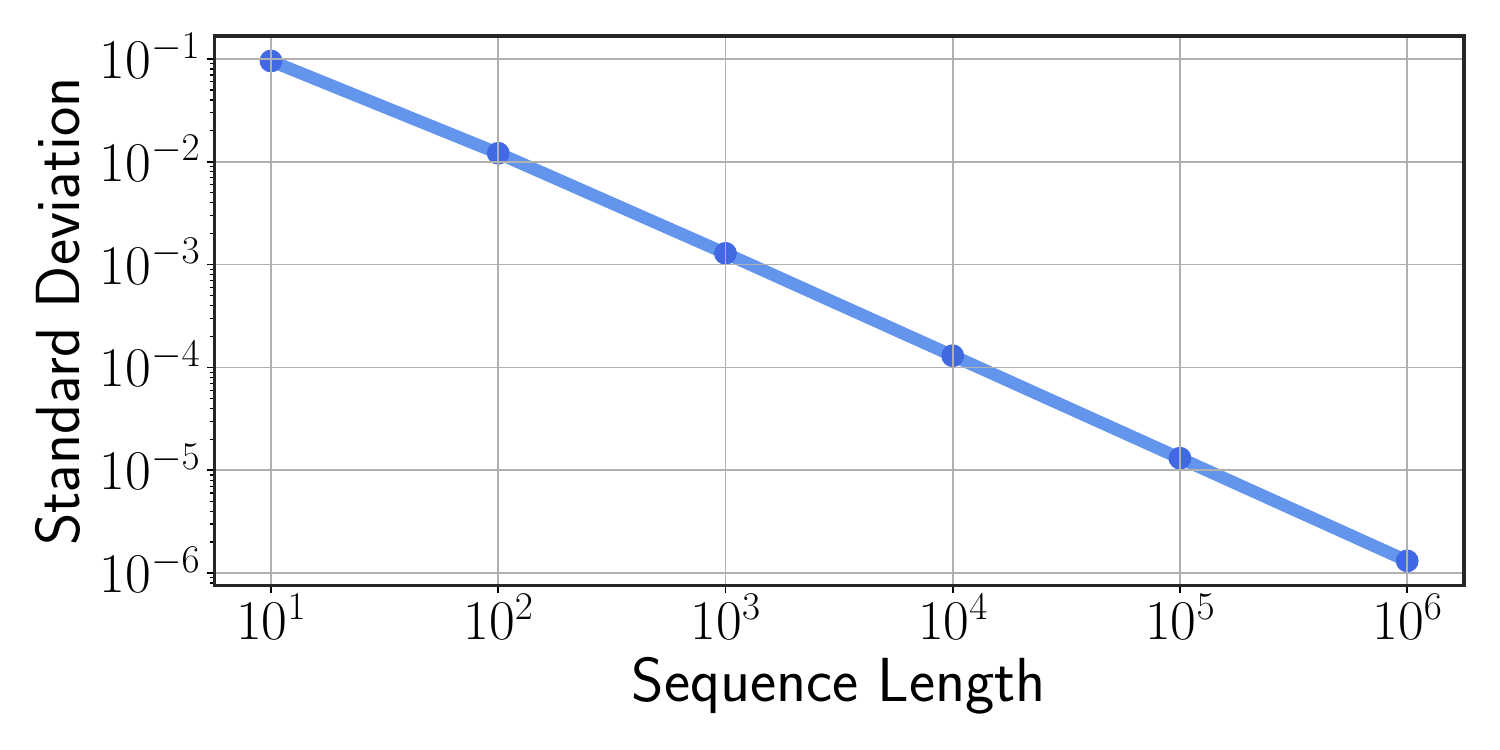}
    \caption{
    Setting the sequence length $N\in\{10,...,10^6\}$, we run the code $100$ trials with Pytorch 2.0.0 and Python 3.9. 
    With the increase of $N$, the softmax operation suffers from the over-smoothness issue.}
    \label{fig:sm-std}
\end{figure}

It has been known that the attention map in~\eqref{eq:att_softmax} suffers from the following two drawbacks:
\begin{itemize}
\item \textbf{Numerical Issues of Softmax.} 
The softmax operation makes the attention maps dense and row-wisely normalized, which suffers from numerical issues when dealing with long sequences. 
In particular, the output of the softmax operation tends to be over-smoothed with the increase in data size.
Given an vector $\bm{x}\in\mathbb{R}^N$, we set $\bm{y}=[y_n]=\text{Softmax}(\bm{x})$, where $y_n$ is the $n$-th element of $\bm{y}$. 
The standard deviation of the elements, i.e., $\frac{1}{N-1}\sum_{n=1}^{N}(y_n - \frac{1}{N}\sum_{n'=1}^{N}y_{n'})$, reduces rapidly when $N$ increases, as shown in Fig.~\ref{fig:sm-std}.
In other words, the elements in $\bm{y}$ become indistinguishable. 
\item \textbf{High Complexity of QKV Architecture.}
As shown in~\eqref{eq:att_softmax}, the computation of an attention map involves a matrix multiplication, whose time complexity is $\mathcal{O}(DN^2)$. 
Moreover, because of using the softmax operation, $\bm{P}(\bm{Q},\bm{K})$ is always a dense matrix no matter whether $\bm{QK}^{\top}$ is sparse or not.
As a result, the space complexity of the attention map is $\mathcal{O}(N^2)$. 
Both the time and space complexity are quadratic to the length of the sequence.
\end{itemize}
Because of the two drawbacks, without any additional data preprocessing, like the LSH in~\cite{kitaev2020reformer} and the subsequence sampling in~\cite{beltagy2020longformer,child2019generating}, the attention maps used in the current MHA mechanism are always over-smoothed and dense when modeling long sequences, which does harm to model performance and computational efficiency severely. 

To overcome the drawbacks, some models apply sparse attention maps, e.g., CosFormer~\cite{zhen2022cosformer} and Longformer~\cite{beltagy2020longformer}, and achieve competitive performance and higher efficiency than the vanilla Transformer, as shown in Fig.~\ref{fig:cmp} and the results reported in~\cite{beltagy2020longformer,gu2021efficiently,zhen2022cosformer,ma2022mega}. 
Besides introducing sparse structures, at the same time, some attempts are made to impose low-rank structures on attention maps, e.g., Local Attention~\cite{tay2021long} and Linear Transformer~\cite{katharopoulos2020transformers}. 
Although these models can also reduce the computational complexity, they seem to harm the performance --- the gaps between the vanilla Transformer and the models using low-rank attention maps are significant on the LRA benchmark, as shown in Fig.~\ref{fig:cmp}. 
In our opinion, a potential reason for this phenomenon is that when $\text{rank}(\bm{P})\ll N$, the rank of the output embeddings $\text{MHA}_{\theta}(\bm{X})$ is likely to be smaller than $N$. 
The low-rank output suffers from a high risk of mode collapse, whose representation power is limited.

In addition, the recent work in~\cite{sander2022sinkformers} shows that in various discriminative learning tasks, the attention maps tend to be doubly stochastic (i.e., $\bm{P}\bm{1}_N=\bm{1}_N$ and $\bm{P}^{\top}\bm{1}_N=\bm{1}_N$) during training.\footnote{See Fig.~2 in~\cite{sander2022sinkformers} for more details.} 
Accordingly, a new variant of Transformer, called Sinkformer~\cite{sander2022sinkformers}, is proposed, which implements the attention maps based on the Sinkhorn-Knopp algorithm~\cite{sinkhorn1967concerning} and makes them doubly stochastic strictly. 
The Sinkformer performs better than the vanilla Transformer in many discriminative learning tasks, e.g., image and text classification.

\textbf{In this study, we aim to replace the attention map in~\eqref{eq:att_softmax} with a new operation, and accordingly, propose a surrogate of the MHA in~\eqref{eq:mha} with better (or at least comparable) model performance and higher computational efficiency.} 
In particular, the drawbacks of the current attention map and its recent advances provide us with valuable insights into the model design. 
On the one hand, to reduce the computational complexity and overcome the over-smoothness issue, it is necessary for us to apply a sparse attention map. 
On the other hand, applying a full-rank and doubly stochastic attention map helps to preserve or improve model performance. 
Therefore, we would like to design an attention map satisfying the above structural constraints (i.e., \textbf{sparse, full-rank, and doubly stochastic}).
In the following subsection, we will show that a simple and novel slicing-sorting operation can implicitly achieve such a structured attention map.

\subsection{Slicing-Sorting for Implicit Structured Attention}

As shown in Table~\ref{tab:cmp}, the key contribution of our Sliceformer is implementing a new attention layer based on an extremely simple \textbf{slicing-sorting} operation, which projects the input linearly to a latent space and sorts each feature, i.e.,
\begin{eqnarray}\label{eq:ss}
    \text{SliceSort}(\bm{X}) := \text{Sort}_{\text{col}}(\underbrace{\bm{X}\bm{W}_V}_{\bm{V}=[\bm{v}_i]}) 
    = \text{Concat}_{\text{col}}(\{\bm{P}_i\bm{v}_i\}_{i=1}^{MD}),
\end{eqnarray}
where $\bm{W}_V\in\mathbb{R}^{d\times MD}$ is the projection matrix,\footnote{Here, we set the number of columns in $\bm{W}_V$ to be $MD$, such that the output has the same shape with the output of MHA.} and $\bm{V}=\bm{XW}_V$. 
Here, we call each column of $\bm{V}$ a ``slice'', denoted as $\bm{v}_i$ for $i=1,...,MD$. 
Each slice corresponds to the projection result of $N$ $d$-dimensional samples in a 1D space. 
As shown in~\eqref{eq:ss}, sorting a slice $\bm{v}_i$ corresponds to the multiplication between a permutation matrix $\bm{P}_i$ and the slice, and accordingly, our slicing-sorting operation concatenates all sorted slices as the output, i.e., $\text{SliceSort}(\bm{X})\in\mathbb{R}^{N\times MD}$. 

Compared to the current MHA architecture, the slicing-sorting operation has several advantages.
\begin{itemize}
    \item \textbf{Implicit Structured Attention Maps.} 
    For each slice $\bm{v}_i$, the sorting step implements its attention map implicitly as the permutation matrix $\bm{P}_i$.
    The permutation matrix naturally satisfies the three desired structural constraints --- it is sparse, full-rank, and doubly stochastic, leading to a competitive alternative to the traditional attention map.
    Instead of applying the same attention map to a group of $\bm{v}_i$'s, we compute a specific permutation matrix for each $\bm{v}_i$. 
    In other words, the number of attention heads in our slicing-sorting operation can be much larger than in the current MHA architecture. 
    \item \textbf{Low Computational Complexity.}
    Given $\bm{V}\in\mathbb{R}^{N\times MD}$, we sort the $MD$ columns, whose time complexity is $\mathcal{O}(MDN\log N)$. 
    On the contrary, the time complexity of the other attention layers (i.e., the complexity per head in Table~\ref{tab:cmp} times the number of heads $M$) is at most comparable to ours.
    With the increase in input sequence length, the time gap between training and inference for the slicing-sorting and other attention layers will become more significant.
    Moreover, instead of generating explicit attention maps with size $\mathcal{O}(N^2)$, we implement permutation matrices implicitly via sorting, whose space complexity is $\mathcal{O}(\log N)$ in general and $\mathcal{O}(N)$ in the worst case. 
    Because of abandoning the QKV architecture, our slicing-sorting has a huge advantage on space complexity compared to the other attention layers.
\end{itemize}

\subsection{Implementations of Sliceformer}\label{ssec:variants}
Replacing the MHA layer with our slicing-sorting operation, we obtain the proposed Sliceformer model, which has fewer parameters and higher computational efficiency than the original Transformer.
The sorting step of the slicing-sorting operation plays a central role in the Sliceformer, which determines the attention map applied to each channel.
Typically, we can implement the sorting step in ascending order (or equivalently in descending order) and apply it to each layer of Sliceformer. 
To further investigate the rationality of our model design, we can apply the following variants of the slicing-sorting operation, leading to different implementations of the Sliceformer.
In particular, given $\bm{V}\in\mathbb{R}^{N\times MD}$, we consider the following two settings for the sorting step.

\begin{itemize}
    \item \textbf{Max-Exchange.} 
    For each $\bm{v}_i$, we find its maximum element and exchange it with the first element, whose computational complexity is $\mathcal{O}(N)$ in time and $\mathcal{O}(1)$ in space. 
    This setting leads to a partial permutation matrix for each $\bm{v}_i$, which only has two non-diagonal elements. 
    As a result, this variant further simplifies the slicing-sorting operation, constructing the attention maps satisfying the structural constraints and with lower computational complexity.
    \item \textbf{Order-Interleave.} We can sort the columns of $\bm{V}$ in different orders and interleave the orders with different frequencies in different layers. 
    Given a Sliceformer with $L$ layers, let $\bm{V}$ be the value matrix in the $n$-th layer. 
    For its $i$-th column, denoted as $\bm{v}_i^{(n)}$, we have
    \begin{eqnarray}\label{eq:haar_sort}
    \begin{aligned}
        \text{Sort}(\bm{v}_i^{(n)})=
        \begin{cases}
            \text{Ascending}(\bm{v}_i^{(n)}), &  \psi_{n}(i)\geq 0,\\
            \text{Descending}(\bm{v}_i^{(n)}), & \psi_{n}(i)< 0,
        \end{cases}
    \end{aligned}
    \end{eqnarray}
    where $\psi_{n}(i)$ is defined as
    \begin{eqnarray}\label{eq:func}
    \begin{aligned}
        \psi_{n}(i)=\sin\Bigl(2^{L-n}\pi\frac{i}{MD}\Bigr),~i=1,...,MD.
    \end{aligned}
    \end{eqnarray}
    By applying different sorting orders, we can increase the diversity of the attention maps across different layers.
\end{itemize}
In the following experiments, we will show that even if applying the max-exchange strategy, our Sliceformer can achieve competitive performance with low complexity. 
The more complicated slicing-sorting operations, i.e., those sorting $\bm{V}$'s in consistent ascending order or interleaved ascending and descending orders, lead to the Sliceformer models outperforming state-of-the-art Transformer-based models.

\section{Experiments}\label{sec:exp}
We demonstrate the effectiveness and efficiency of our Sliceformer in discriminative tasks through comprehensive comparative and analytic experiments. 
In particular, we first compare our Sliceformer to Transformer and its representative variants on the well-known Long Range Arena (LRA) benchmark~\cite{tay2021long} and empirically verify the rationality of our slicing-sorting operation in long sequence classification tasks. 
Then, we implement ViT~\cite{dosovitskiy2021an} by our Sliceformer and test its performance and permutation-invariance in image classification tasks. 
Finally, we explore the applications of our Sliceformer in other domains, including text classification and graph classification, demonstrating the universal applicability of our slicing-sorting attention layer.
In addition, we further analyze the singular spectrum achieved by our slicing-sorting operation and show that our Sliceformer has a lower risk of mode collapse empirically, which provides a potential explanation for its encouraging performance. 
For convenience, we denote the Sliceformers applying the ``Ascending Order'', ``Order-Interleave'', and ``Max-Exchange'' strategies as \textbf{Sliceformer$_{\text{ascend}}$}, \textbf{Sliceformer$_{\text{interleave}}$}, and \textbf{Sliceformer$_{\text{max}}$}, respectively.

\subsection{Long Range Arena Benchmark}
Long Range Arena (LRA) is a benchmark designed to evaluate models for long sequence scenarios~\cite{tay2021long}, which consists of six discriminative learning tasks, including ListOps~\cite{nangia2018listops}, byte-level text classification~\cite{maas2011learning}, byte-level document retrieval~\cite{radev2013acl}, and three sequentialized image classification tasks, i.e., CIFAR-10~\cite{krizhevsky2009learning}, Pathfinder~\cite{linsley2018learning},\footnote{Pathfinder is an image classification task: given a set of gray-level images, each of which plots two points and several curves, the model aims to recognize whether there exists a path connecting the points in each image.} and Pathfinder-X (a longer and more challenging version of Pathfinder). 
Each image is formulated as a long sequence of pixels in the three image classification tasks.
We test our Sliceformer on the LRA benchmark and compare it to state-of-the-art Transformer-based models on both prediction accuracy and computational efficiency. 
In each task, our Sliceformer is trained to represent each input sequence through the embedding in the ``CLS'' position. 
For a fair comparison, we implement all the models based on JAX~\cite{bradbury2018jax} and strictly follow the benchmark's default data processing and experimental design.
In each task, our Sliceformer is the same as the other models on the number of layers and the dimension of hidden variables. 
Hence, its model parameters are fewer because of abandoning the QKV architecture. 
All the models are trained on four NVIDIA 3090 GPUs. 

\begin{table}[t]
  \centering
  \caption{The comparison for various models on the LRA benchmark. 
  For each model, we record its classification accuracy (\%) in each task and the average performance.
  ``FAIL'' means the training process fails to converge.
  In each column, we bold the best result and underline the second best one.}
  % \small{
  \tabcolsep=2pt
    \begin{tabular}{l|ccccccc}
    \toprule
    \multicolumn{1}{c|}{Model} & ListOps & Text  & Retrieval & Image & Path  & Path-X & Avg.\\
    \midrule
    Transformer~\cite{vaswani2017attention} & 36.37 & 64.27 & 57.46 & 42.44 & 71.40 & FAIL  & 54.39 \\
    \midrule
    Local Att.~\cite{tay2021long} & 15.82 & 52.98 & 53.39 & 41.46 & 66.63 & FAIL  & 46.06 \\
    Linear Trans.~\cite{katharopoulos2020transformers} & 16.13 & \textbf{65.90} & 53.09 & 42.34 & 75.30 & FAIL  & 50.55 \\
    Reformer~\cite{kitaev2020reformer}  & 37.27 & 56.10 & 53.40 & 38.07 & 68.50 & FAIL  & 50.67 \\
    Sinkformer~\cite{sander2022sinkformers} & 30.70 & 64.03 &  55.45 & 41.08 & 64.65 & FAIL & 51.18 \\
    SparseTrans~\cite{child2019generating} & 17.07 & 63.58 & 59.59& 44.24 & 71.71 & FAIL  & 51.24 \\
    SinkhornTrans~\cite{tay2020sparse} & 33.67 & 61.20 & 53.83 & 41.23 & 67.45 & FAIL  & 51.29 \\
    Linformer~\cite{wang2020linformer} & 35.70 & 53.94 & 52.27 & 38.56 & 76.34 & FAIL  & 51.36 \\
    Performer~\cite{choromanski2021rethinking} & 18.01 & \underline{65.40} & 53.82 & 42.77 & 77.05 & FAIL  & 51.41 \\
    Synthesizer~\cite{tay2021synthesizer} & 36.99 & 61.68 & 54.67 & 41.61 & 69.45 & FAIL  & 52.88 \\
    Longformer~\cite{beltagy2020longformer} & 35.63 & 62.85 & 56.89 & 42.22 & 69.71 & FAIL  & 53.46 \\
    BigBird~\cite{zaheer2020big} & 36.05 & 64.02 & 59.29 & 40.83 & 74.87 & FAIL  & 55.01 \\
    Cosformer~\cite{zhen2022cosformer} & \textbf{37.90} & 63.41 & 61.36 & 43.17 & 70.33 & FAIL  & 55.23 \\
    \midrule
    Sliceformer$_{\text{max}}$ & 37.00 & 62.90 & 59.00 & 40.48 & 75.42 & 
    FAIL  & 54.96 \\
    Sliceformer$_{\text{interleave}}$ & 37.30 & 64.25 & \underline{61.97} & \underline{45.88} & \underline{81.98} & FAIL  & \underline{58.28} \\
    Sliceformer$_{\text{ascend}}$ & \underline{37.65} &  64.60     &  \textbf{62.23}     &   \textbf{48.02}    &  \textbf{82.04}      &   FAIL  & \textbf{58.91}  \\
    % & Half-Half  & 37.45 & 64.25 & 61.95 & 45.80 & 81.38 & FAIL  & 58.17 \\
    % & Multi-Permutation ($K=2$) & 37.25 & 63.70 & 58.42 & 47.37 & 81.04 & FAIL & 57.56\\
    % & Multi-Permutation ($K=4$) & 37.35 & 63.80 & 57.16 & 46.92 & 74.06 & FAIL & 55.86\\
    % & Shuffle & 17.85 & 50.49 & 50.87 & 10.00 & 49.74 & FAIL  & 35.79 \\
    % \midrule
    % CCNN~\cite{romero2022towards} & 43.60 & 84.08 & FAIL & \underline{88.90} & 91.51 & FAIL & 68.02 \\
    % Mega~\cite{ma2022mega} & \textbf{63.14} & \textbf{90.43} & \underline{91.25} & \textbf{90.44} & \textbf{96.01} & \underline{97.98} & \textbf{88.21} \\
    % S4~\cite{gu2021efficiently} & 59.60 & 86.82 & 90.90 & 88.65 & 94.20 & 96.35 & 86.09  \\
    % S5~\cite{smith2022simplified} & \underline{62.15} & \underline{89.31} & \textbf{91.40} & 88.00 & \underline{95.33} & \textbf{98.58} & \underline{87.46}  \\
    \bottomrule
    \end{tabular}%
  \label{tab:lra_res}%
  % }
\end{table}%

As shown in Table~\ref{tab:lra_res}, when applying the Ascending Order or the Order-Interleave strategy, our Sliceformer performs the best in three of the six tasks and achieves the highest average accuracy. 
Especially in the challenging image classification tasks on CIFAR-10 and Pathfinder, our Sliceformer outperforms the other models significantly, improving the classification accuracy by at least four percentage points.
Even if applying the Max-Exchange strategy, the overall performance of Sliceformer can be comparable to the state-of-the-art models, e.g., BigBird~\cite{zaheer2020big} and Cosformer~\cite{zhen2022cosformer}. 

\begin{table*}[t]
  \centering
  \caption{The comparison for various models on their computational efficiency. 
  ``OOM'' means the training process suffers from the out-of-memory issue.
  In each column, we bold the best result and underline the second best one.}
  % \small{
    \begin{tabular}{l|cccc|cccc|cccc}
    \toprule
    \multirow{2}{*}{Model}  & \multicolumn{4}{c|}{Training speed (steps per second)} & \multicolumn{4}{c|}{Inference speed (steps per second)} & \multicolumn{4}{c}{Peak Memory Usage (GB)} \\
     & 1K    & 2K    & 3K    & 4K & 1K    & 2K    & 3K    & 4K   &  1K    & 2K    & 3K    & 4K \\
    \midrule
    Transformer~\cite{vaswani2017attention} & 27.49 & 9.45  & 4.73  & OOM & 41.54 & 32.90 & 16.09 & OOM & 11.64 & 32.45 & 65.67 &  OOM \\
    \midrule
    Local Attention~\cite{tay2021long} & 31.41 & 25.47 & 18.06 & 13.80 & 41.24 & 41.34 & 42.35 & \underline{42.27} &  6.23  & 9.24  & 12.26 & 15.27 \\
    Linear Trans.~\cite{katharopoulos2020transformers} & 31.35 & 25.79 & 17.07 & 12.32 & 41.98 & \underline{42.51} & \underline{42.39} & 42.12 & 6.50  & 9.84  & 13.18 & 16.52 \\
    Reformer~\cite{kitaev2020reformer} & 31.55 & 22.00 & 13.42 & 8.84 & \underline{42.80} & 42.25 & 41.65 & 34.25  &  6.91  & 12.22 & 19.54 & 28.88 \\
    Sinkformer~\cite{sander2022sinkformers} & 18.72 & 5.82  & 2.86  & OOM & 41.74 & 20.49 & 9.12 & OOM  & 14.13 & 41.58 & 82.53 & OOM \\
    SparseTrans~\cite{child2019generating} & 27.39 & 9.47  & 4.72  & OOM & 41.16 & 40.45 & 41.37 & 42.12 & 11.64 & 32.45 & 65.71 & OOM \\
    SinkhornTrans~\cite{tay2020sparse} & 29.88 & 22.00 & 15.66 & 11.70 & 41.16 & 40.45 & 41.37 & 42.12 &  6.70  & 10.23 & 13.78 & 17.32 \\
    Linformer~\cite{wang2020linformer} & \underline{28.47} & \underline{28.08} & \underline{19.08} & \underline{14.55} & 41.72 & 42.42 & 42.23 & 42.12 &  \underline{5.95}  & \underline{8.82}  & \underline{11.65} & \underline{14.47} \\
    Performer~\cite{choromanski2021rethinking} & 28.84 & 26.67 & 18.97 & 14.10 & 41.50 & 42.31 & 41.87 & 41.69 & 6.25  & 9.35  & 12.45 & 15.54 \\
    Synthesizer~\cite{tay2021synthesizer} & 19.57 & 6.27  & 3.01  & OOM & 41.84 & 27.44 & 13.21 &  OOM &  12.77 & 37.53 & 75.23 & OOM  \\
    Longformer~\cite{beltagy2020longformer} & 17.56 & 5.67  & 2.74  &  OOM & 40.91 & 32.52 & 15.79 & OOM & 13.00 & 37.07 & 75.46 & OOM\\
    BigBird~\cite{zaheer2020big} & 27.27 & 14.34 & 9.76  & 7.21 & 40.84 & 41.50 & 34.85 & 26.02   & 9.59  & 16.53 & 23.16 & 30.07 \\    
    Cosformer~\cite{zhen2022cosformer} & 28.42 & 26.13 & 17.56 & 12.58 & 41.62 & 40.78 & 40.56 & 40.50 & 6.36 & 9.68 & 13.36 & 16.95 \\
    \midrule
    % Sliceformer$_{\text{max}}$ & 30.99 & 30.73 & 23.28 & 17.19 & 40.88 & 40.66 & 40.40 & 40.15 & 7.47 & 11.73 & 11.77 & 20.26\\
    Sliceformer$_{\text{ascend}}$ & \textbf{32.79} & \textbf{32.26} & \textbf{21.79} & \textbf{16.25} & \textbf{43.73} & \textbf{43.72} & \textbf{43.29} & \textbf{42.89}   & \textbf{5.61}  & \textbf{8.12}  & \textbf{10.64} & \textbf{13.14} \\
    \bottomrule
    \end{tabular}%
    % }
  \label{tab:lra_effi}%
\end{table*}%

Table~\ref{tab:lra_effi} further compares the models' training speed, inference speed, and peak memory usage when dealing with sequences ranging from 1K to 4K.
Note that, according to Table~\ref{tab:lra_res}, both Sliceformer$_{\text{ascend}}$ and Sliceformer$_{\text{interleave}}$ outperform other models. 
They apply the sorting step and thus have the same complexity. 
Therefore, we mainly focus on the efficiency analysis of Sliceformer$_{\text{interleave}}$ in Table~\ref{tab:lra_effi}.
We can find that the most efficient model among the baselines is Linformer~\cite{wang2020linformer}, but its average accuracy on LRA is merely 51.36\%. 
Our Sliceformer$_{\text{ascend}}$ is more efficient than the other Transformer-based models, executing more training and inference steps per second and occupying less memory.
Its advantage in computational efficiency becomes even more significant with the increase of the sequence length.  

The results in Tables~\ref{tab:lra_res} and~\ref{tab:lra_effi} have also been illustrated in Figure~\ref{fig:cmp}.
In summary, our Sliceformer$_{\text{ascend}}$ and Sliceformer$_{\text{interleave}}$ achieve a trade-off between model performance and computational efficiency.
When classifying long sequences, they obtain comparable or superior accuracy with significant improvements in runtime compared to the Transformer and its variants.

\begin{table*}[t]
    \centering
  \caption{The comparison for our Sliceformer and ViT on the number of parameters ($\times 10^6$) and classification accuracy (\%). In each task, we bold the best result and underline the second best one.\label{tab:vit}}
  % \small{
  \tabcolsep=3pt
    \begin{tabular}{l|cc|cc|cc|cc|cc}
    \toprule
    Data &\multicolumn{2}{c|}{Dogs vs. Cats} &\multicolumn{2}{c|}{MNIST}&\multicolumn{2}{c|}{CIFAR-10} &\multicolumn{2}{c|}{CIFAR-100} &\multicolumn{2}{c}{Tiny-ImageNet} \\
    Metric & Model Size & Top-1 Acc & Model Size & Top-1 Acc & Model Size & Top-1 Acc& Model Size & Top-1 Acc& Model Size & Top-1 Acc\\
    \midrule
    ViT & 1.90 & 79.03 & 9.60 & 98.78 & 9.60 & 80.98 & 9.65 & 53.99 & 22.05 & \textbf{52.74} \\
    Sliceformer$_{\text{ascend}}$ & \textbf{1.11} & \underline{79.71} & \textbf{6.50} & \underline{98.81} & \textbf{6.46} & \underline{82.16} & \textbf{6.50} & \underline{54.24} & \textbf{18.50} & 51.77 \\
    Sliceformer$_{\text{interleave}}$ & \textbf{1.11} & \textbf{79.87} & \textbf{6.50} & \textbf{99.00} & \textbf{6.46} & \textbf{83.54} & \textbf{6.50} & \textbf{54.70} & \textbf{18.50} & \underline{52.40} \\
    \bottomrule
    \end{tabular}
    % }
\end{table*}

\subsection{Testing on Other Discriminative Tasks}
Besides modeling long sequences, we test our Sliceformer models in other applications, demonstrating their universality. 
In particular, focusing on Sliceformer$_{\text{ascend}}$ and Sliceformer$_{\text{interleave}}$, we apply them to text classification, image classification, and molecular property prediction tasks.

\subsubsection{Image Classification}
\begin{figure}[t]
  \centering
  % \subfigure[Permutation-invariance test]{
  % \includegraphics[height=4.3cm]{figures/MAE.pdf}\label{fig:cos}
  % }
  % \subfigure[Convergence of training process]{
  \includegraphics[height=4.5cm]{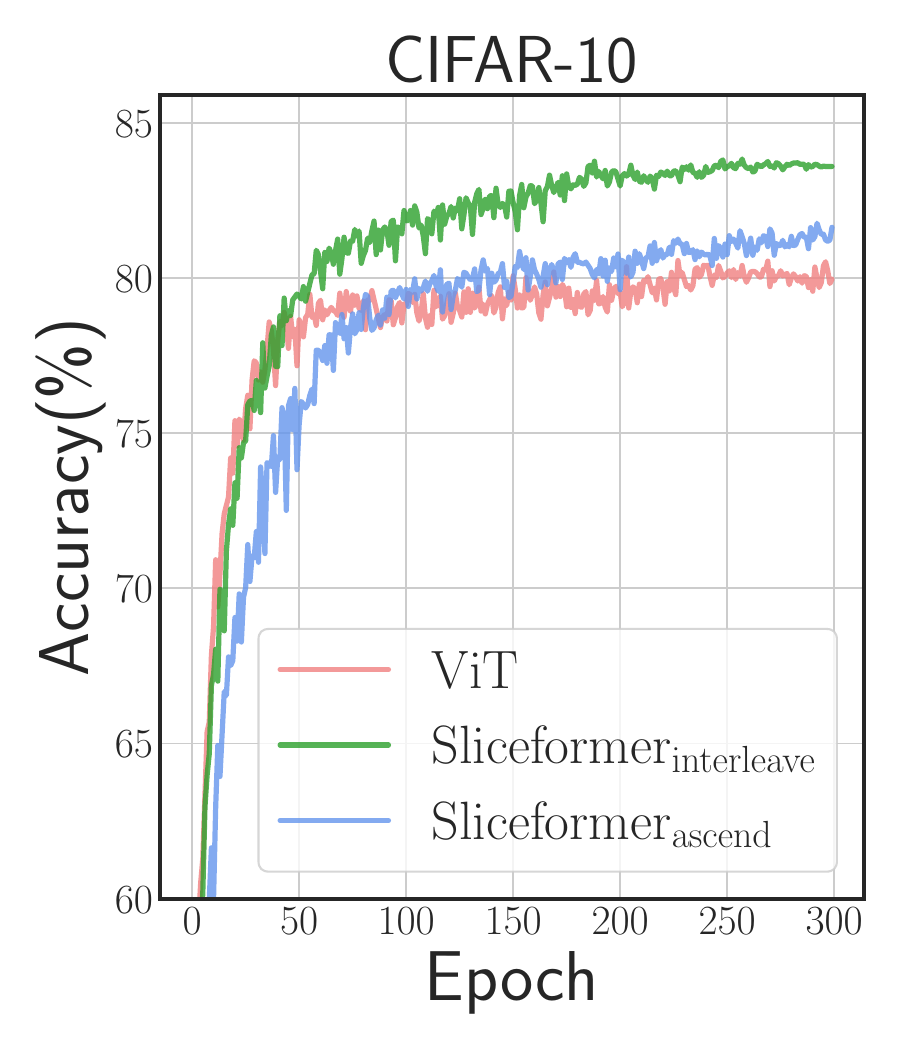}
  \includegraphics[height=4.5cm]{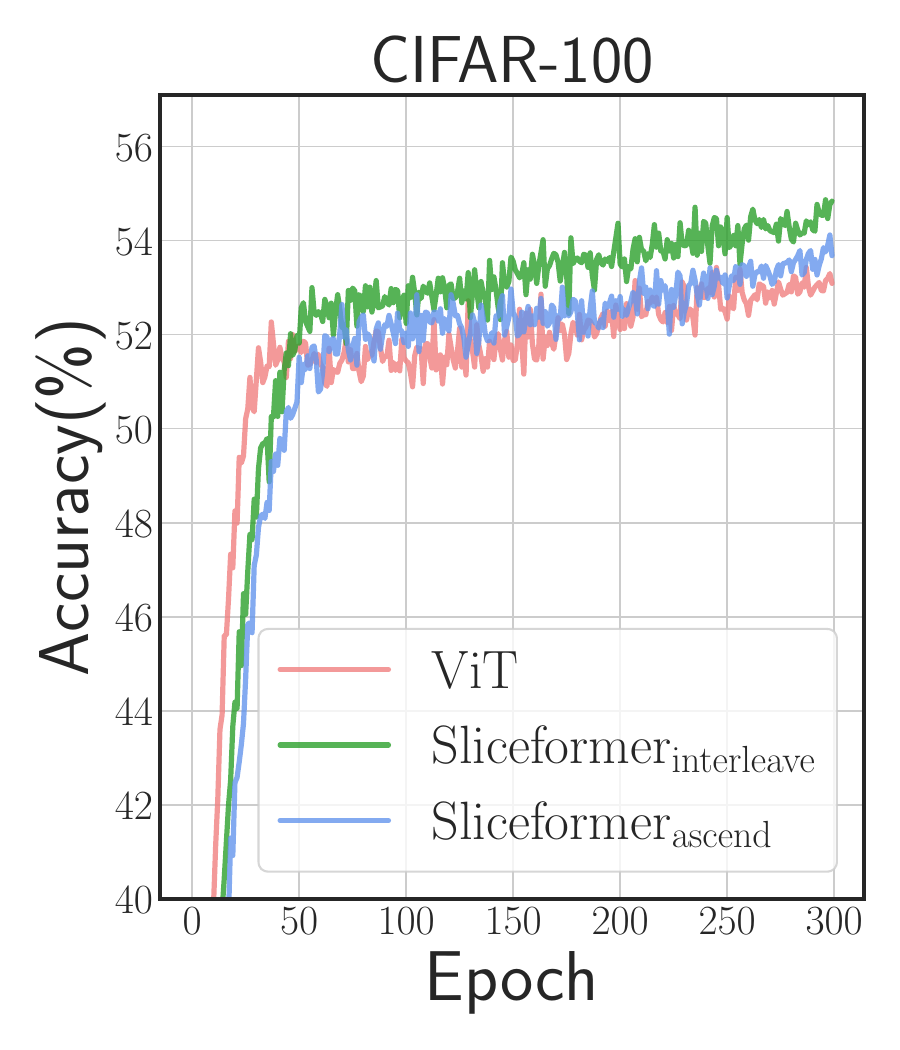}
  % }
  \caption{
  % (a) The impact of the order of pixel sequence on the numerical permutation-invariance of model.
  The comparison for our Sliceformers and ViT on their training convergence.}
  \label{fig:convergence}
\end{figure}

Like ViT~\cite{dosovitskiy2021an}, we can treat images as patch sequences and apply our Sliceformer to achieve image classification.
In particular, by replacing the MHA layers of ViT with our slicing-sorting operation, we obtain the ``Vision Sliceformer'' accordingly.
We test the Sliceformer on five image datasets, including Dogs vs. Cats~\footnote{https://www.kaggle.com/c/dogs-vs-cats/data}, MNIST, CIFAR-10, CIFAR-100~\cite{krizhevsky2009learning}, and Tiny-ImageNet~\cite{le2015tiny}. 
For each dataset, we compare Sliceformer$_{\text{ascend}}$ and Sliceformer$_{\text{interleave}}$ to ViT on their model size and classification accuracy. 
For these three models, we set the number of layers and the dimension of each layer to be the same and train them from scratch.
Table~\ref{tab:vit} compares the classification accuracy achieved by the three models, and Fig.~\ref{fig:convergence} further illustrates the convergence of the model performance with the increase of training epochs on CIFAR-10 and CIFAR-100, respectively. 
The results show that our Sliceformers outperform ViT consistently on both model size and classification accuracy.
Significantly, the Sliceformer$_{\text{interleave}}$ achieves the best performance on four of the five datasets with fewer parameters.

\subsubsection{Text Classification}
To further evaluate the language modeling capability of Sliceformer, we test it on the IMDB dataset~\cite{yang2019xlnet} and compare it to the Transformer. 
As shown in Table~\ref{tab:text}, Sliceformer consistently outperforms the Transformer on the IMDB dataset, and at the same time, its model size is smaller than that of the Transformer. 
This result further demonstrates the superiority of the Sliceformer to the Transformer.

\begin{table}[t]
    \centering
  \caption{The comparison the number of parameters ($\times 10^6$) and text classification accuracy (\%).\label{tab:text}}
  % \small{
  % \tabcolsep=4pt
    \begin{tabular}{l|cc}
    \toprule
    Data &\multicolumn{2}{c}{IMDB}\\
    Metric & Model Size & Top-1 Acc. \\
    \midrule
    Transformer & 8.84 & 83.05 \\
    Sliceformer$_{\text{ascend}}$ & \textbf{8.05} & \textbf{84.91} \\
    Sliceformer$_{\text{interleave}}$ & \textbf{8.05} & \underline{84.55} \\
    \bottomrule
    \end{tabular}
    % }
\end{table}

\subsubsection{Molecular Property Prediction}
Our Sliceformer is also applicable to graph-structured data like molecules.
In this study, we introduce the slicing-sorting operation to Graphormer~\cite{ying2021transformers} and test its impacts on molecular property prediction. 
In particular, the attention head of Graphormer applies a modified QKV architecture, which is formulated as $\text{Softmax}(\bm{S}+\bm{E}+\frac{1}{\sqrt{D}}\bm{QK}^{\top})\bm{V}$. 
Here, $\bm{S}$ and $\bm{E}$ are learnable embedding matrices encoding spatial positions and edge information, respectively. 
Applying the slicing-sorting operation, we design a simplified attention layer as $\text{Sort}_{\text{col}}(\text{Softmax}(\bm{S}+\bm{E})\bm{V})$, where the query and key matrices are ignored. 
Accordingly, the number of trainable parameters dramatically reduces by around 30\%.
Replacing the attention layer of Graphormer with this layer leads to a Sliceformer for molecular data.
We apply the PCQM4M-LSC dataset~\cite{hu2021ogb} for training and testing the models. 
The experimental results in Table~\ref{tab:graphormer} show that applying the slicing-sort operation leads to a simplified model with a smaller size, whose performance is comparable to Graphormer.

\begin{table}[t]
  \centering
  \caption{The comparison on the number of parameters ($\times 10^6$) and model performance in molecular property prediction.\label{tab:graphormer}}
  % \small{
  % \tabcolsep=2pt
    \begin{tabular}{l|cc}
    \toprule
    Data &\multicolumn{2}{c}{PCQM4M-LSC} \\
    Metric & Model Size & MAE \\
    \midrule
    Graphormer & 47.09 & \textbf{0.1287} \\
    Sliceformer$_{\text{ascend}}$ & \textbf{32.91} & \underline{0.1308} \\
    Sliceformer$_{\text{interleave}}$ & \textbf{32.91} & 0.1342 \\
    \bottomrule
    \end{tabular}
    % }
\end{table}

\subsection{Empirical Evidence of Model Rationality}
\begin{figure}[t]
  \centering
  \subfigure[ViT]{
  \includegraphics[height=2.6cm]{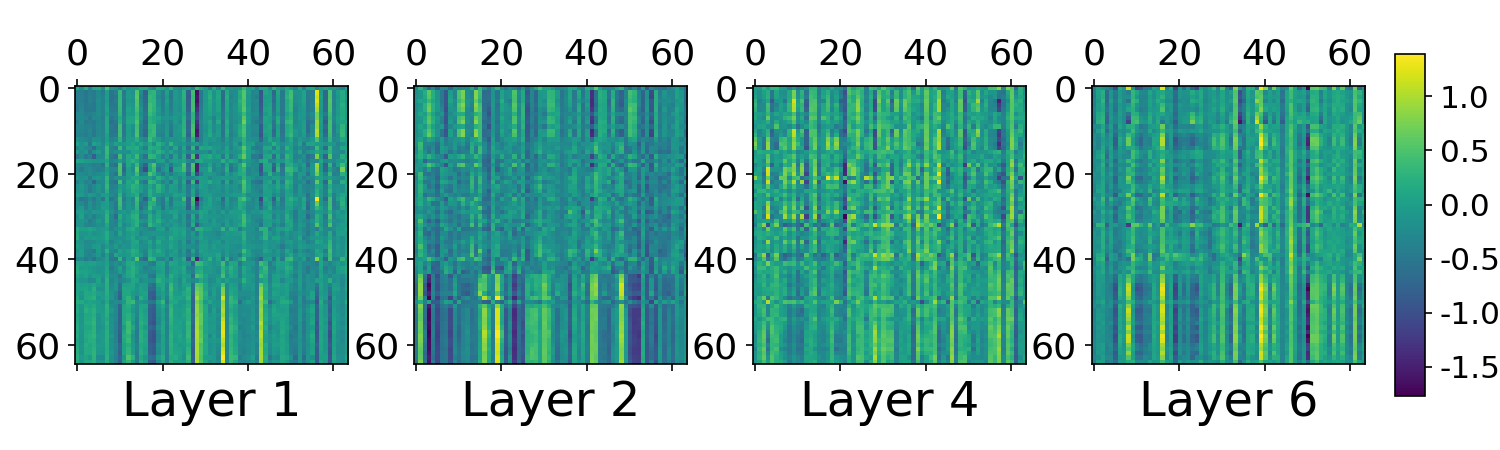}\label{fig:ViT_PxV}
  }
  % \subfigure[Sliceformer$_{\text{max}}$]{
  % \includegraphics[height=2.3cm]{figures/max_PxV2.png}\label{fig:s_max_PxV}
  % }
  \subfigure[Sliceformer$_{\text{ascend}}$]{
  \includegraphics[height=2.6cm]{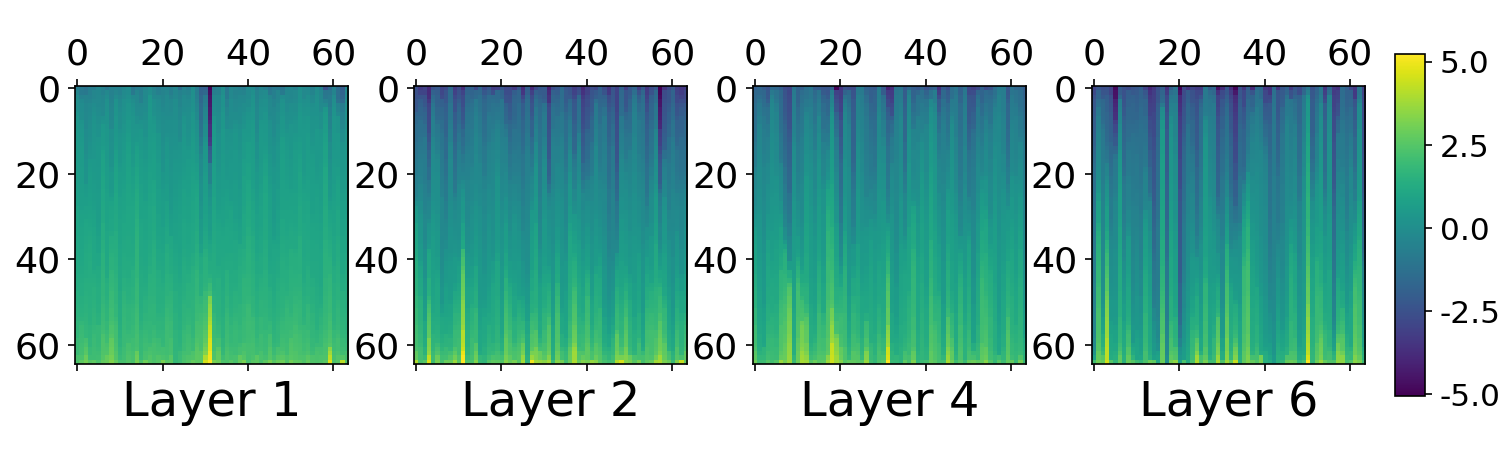}\label{fig:sliceformer_PxV}
  }
  \subfigure[Sliceformer$_{\text{interleave}}$]{
  \includegraphics[height=2.6cm]{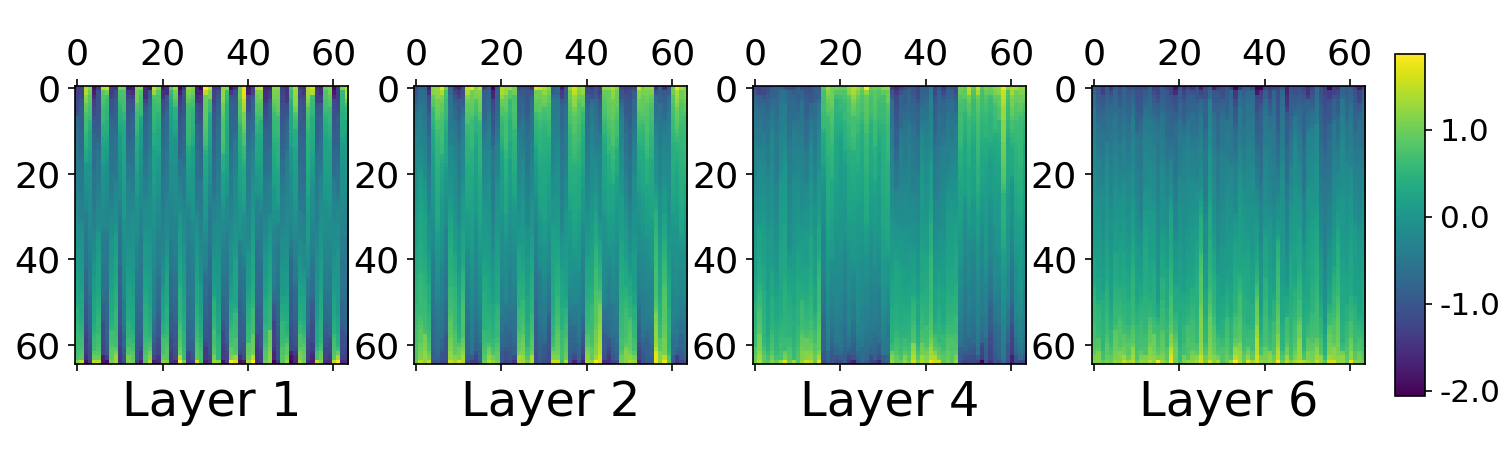}\label{fig:sliceformer_haar_PxV}
  }
  \subfigure[Comparisons on singular spectrum]{
  \includegraphics[height=2.6cm]{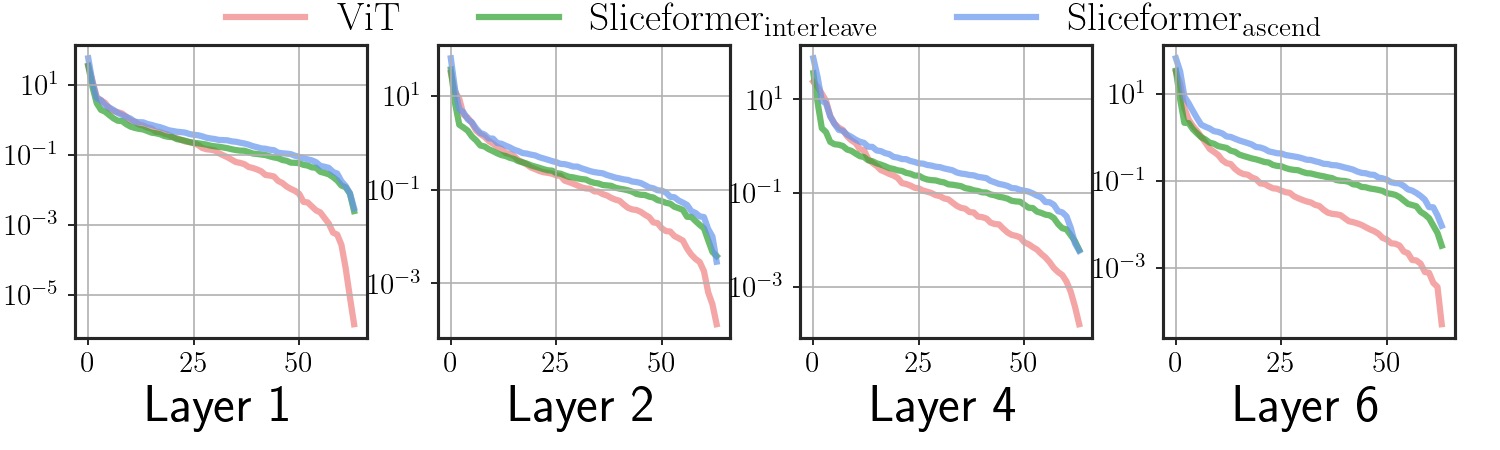}\label{fig:spectrum}
  }
  \caption{
  (a-d) Illustrations of the $\text{MHA}_{\theta}(\bm{X})$ generated by ViT's layers and the $\text{SliceSort}(\bm{X})$ generated by our Sliceformers' layers. 
  Each figure corresponds to the heatmap of the output, whose rows indicate the patches in a sequence and columns indicate the feature channels.
  (e) The comparisons for various models on the singular spectrum of the outputs in different layers.\label{fig:visual}
  }
\end{figure}

After training a ViT and our Sliceformer on CIFAR-10, we visualize the output matrices of their attention layers (i.e., $\text{MHA}_{\theta}(\bm{X})$ and $\text{SliceSort}(\bm{X})$) and analyze their singular spectrum in Fig.~\ref{fig:visual}.
For ViT, the output matrices of its attention layers seem to have low-rank structures, verified by the singular spectrum shown in Fig.~\ref{fig:spectrum} --- the singular values of the output matrices decay rapidly. 
On the contrary, the output matrices obtained by our Sliceformers have sorted columns, as shown in Figs.~\ref{fig:sliceformer_PxV} and~\ref{fig:sliceformer_haar_PxV}. 
Moreover, according to the singular spectrum shown in Fig.~\ref{fig:spectrum}, the singular values of the Sliceformers' output matrices decay much more slowly than those of ViT. 
This phenomenon provides an empirical explanation for the rationality of Sliceformer.
In particular, the slow-decay spectrum indicates that our Sliceformer suppresses the risk of mode collapse when representing data.

\section{Discussion and Conclusion}\label{sec:con}
We have proposed a new data representation model called Sliceformer. 
By replacing the traditional MHA mechanism with a simple slicing-sorting operation, our Sliceformer overcomes the numerical drawbacks of the current Transformers and achieves encouraging performance in various discriminative tasks. 
Our work provides a new perspective to the design of the attention layer, i.e., implementing attention maps through simple algorithmic operations has the potential to achieve low computational complexity and good numerical performance. 

\textbf{Current Limitations and Future Work.} 
The main drawback of Sliceformer is its limited model capacity.
In particular, although satisfying the structural constraints, the implicit attention map in our model is merely a permutation matrix. 
Thus, its representation power is not as good as the softmax-based attention in~\eqref{eq:att_softmax}. 
As a result, when testing on the full-sized ImageNet classification task, the top-1 accuracy of Sliceformer is merely 64.77\%.
To solve this problem, in the future, we would like to develop a differentiable and learnable slicing-sorting operation, enhancing its model capacity by introducing more parameters. 
Additionally, the performance of Sliceformer in generative learning tasks has not been investigated yet. 
We want to improve the model architecture further and make the model applicable to generative learning tasks.
In the aspect of application, we plan to apply Sliceformer to represent structured data like point clouds and meshes.

\bibliographystyle{IEEEtran}
\bibliography{refs}

\vspace{-3mm}
\begin{IEEEbiography}[{\includegraphics[width=1in,height=1.25in,clip,keepaspectratio]{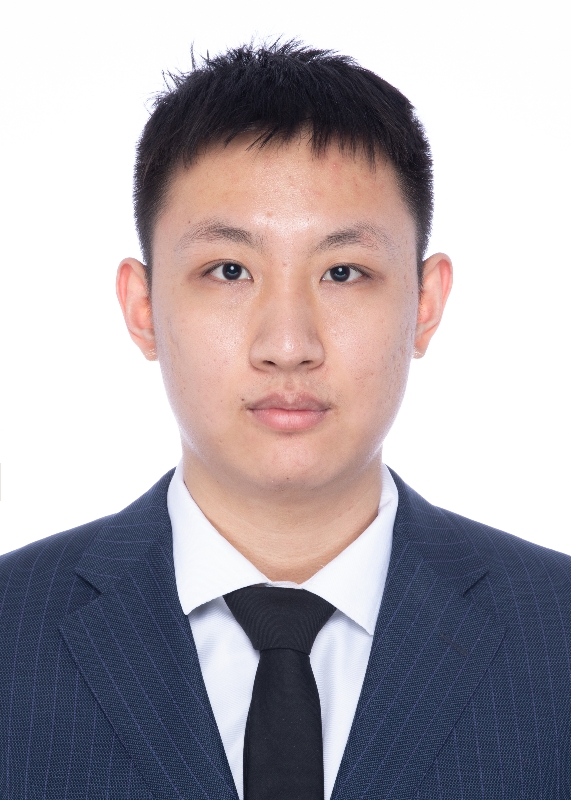}}]{Shen Yuan}
is a third-year Ph.D. candidate, under the supervision of Associate Professor Hongteng Xu, at Gaoling School of Artificial Intelligence, Renmin University of China. 
From February to August 2022, he interned at Tencent AI Lab, supervised by Prof. Lanqing Li and Prof. Peilin Zhao. 
He obtained his undergraduate degree from the School of Computer Science and Engineering, University of Electronic Science and Technology of China.
His research focus is structured data-oriented machine learning and applications. 
His research interests lie in optimal transport, graph modeling, large language models, and AI for drug discovery.
\end{IEEEbiography}
\vspace{-6mm}
\begin{IEEEbiography}[{\includegraphics[width=1in,height=1.25in,clip,keepaspectratio]{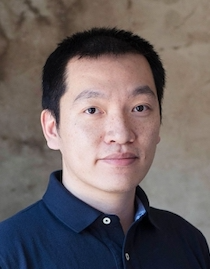}}]{Hongteng Xu}
is an Associate Professor in the Gaoling School of Artificial Intelligence, Renmin University of China. 
From 2018 to 2020, he was a senior research scientist in Infinia ML Inc. 
In the same time period, he is a visiting faculty member in the Department of Electrical and Computer Engineering, Duke University. 
He received his Ph.D. from the School of Electrical and Computer Engineering at Georgia Institute of Technology (Georgia Tech) in 2017. 
His research interests include machine learning and its applications, especially optimal transport theory, sequential data modeling and analysis, deep learning techniques, and their applications in computer vision and data mining.
\end{IEEEbiography}

\end{document}